\documentclass{article} 
\usepackage[final,nonatbib]{nips_2016} 
\usepackage{times}
\usepackage{hyperref}
\usepackage{url}

\usepackage[letterpaper]{geometry}

\usepackage{amsmath,amsfonts,amsbsy}
\usepackage{latexsym}
\usepackage{graphicx}
\usepackage{algorithm,algorithmic}
\usepackage{multicol,multirow}
\usepackage{subfigure}
\usepackage{url}
\usepackage{color}
\usepackage[normalem]{ulem}
\usepackage{caption}

\usepackage{authblk}

\newif\ifemdnn
\emdnntrue

\title{{PVANet}: Lightweight Deep Neural Networks for Real-time Object Detection}
\author{
Sanghoon Hong\thanks{Corresponding author}, Byungseok Roh, Kye-Hyeon Kim\thanks{Current address: kye-hyeon.kim@sktbrain.com}, Yeongjae Cheon, and Minje Park\\
Intel Imaging and Camera Technology, Seoul, Korea\\
\texttt{\{sanghoon.hong, peter.roh, kye-hyeon.kim,}\\
\texttt{yeongjae.cheon, minje.park\}@intel.com} \\
}

\begin{document}

\maketitle

\setcounter{footnote}{0}

\begin{abstract}
\ifemdnn
\fi
In object detection, reducing computational cost is as important as improving accuracy for most practical usages. This paper proposes a novel network structure, which is an order of magnitude lighter than other state-of-the-art networks while maintaining the accuracy. Based on the basic principle of more layers with less channels, this new deep neural network minimizes its redundancy by adopting recent innovations including C.ReLU and Inception structure. We also show that this network can be trained efficiently to achieve solid results on well-known object detection benchmarks: 84.9\% and 84.2\%mAP on VOC2007 and VOC2012 while the required compute is less than 10\% of the recent ResNet-101.
\end{abstract}

\section{Introduction}

Convolutional neural networks (CNNs) have made impressive improvements in object detection for several years.
Thanks to many innovative works, recent object detection algorithms have reached accuracies acceptable for commercialization in a broad range of markets like automotive and surveillance. However, in terms of detection speed, even the best algorithms are still suffering from heavy computational cost. Although recent reports on network compression and quantization shows promising results, it is still important to reduce the computational cost in the network design stage.

The successes in network compression \cite{kim2015compression} and decomposition of convolution kernels \cite{ioannou2015training, iandola2016squeezenet} imply that present network architectures are highly redundant. Therefore, reducing these redundancies is a straightforward approach in reducing the computational cost.

\ifemdnn
This paper presents a lightweight network architecture for object detection, named {\sc PVANet}\footnote{The code and models are available at \url{https://github.com/sanghoon/pva-faster-rcnn}}, which achieves state-of-the-art detection accuracy in real-time. Based on the basic principle of ``smaller number of output channels with more layers'', we adopt C.ReLU\cite{ShangW2016icml} in the initial layers and Inception structure\cite{SzegedyC2015cvpr} in the latter part of the network. Multi-scale feature concatenation\cite{KongT2016cvpr} is also applied to maximize the multi-scale nature of object detection tasks.

We also show that our thin but deep network can be trained effectively with batch normalization \cite{SzegedyC2015icml}, residual connections \cite{HeK2016cvpr}, and our own learning rate scheduling based on plateau detection.
\fi

In the remaining sections, we describe the structure of PVANet as a feature extraction network (Section \ref{subsec:feature_extraction}) and a detection network (Section \ref{subsec:frcnn}). Then, we present experimental results on ImageNet 2012 classification, VOC-2007 and VOC-2012 benchmarks with detailed training and testing methodologies (Section \ref{sec:experiments}).

\section{PVANet}
\label{sec:pvanet}

\subsection{Feature extraction network}
\label{subsec:feature_extraction}


\begin{figure}[t]
\centering
\hspace{-0.5cm}
\begin{minipage}{0.4\textwidth}
    \centering
    \ifemdnn
    \includegraphics[height=0.12\paperheight]{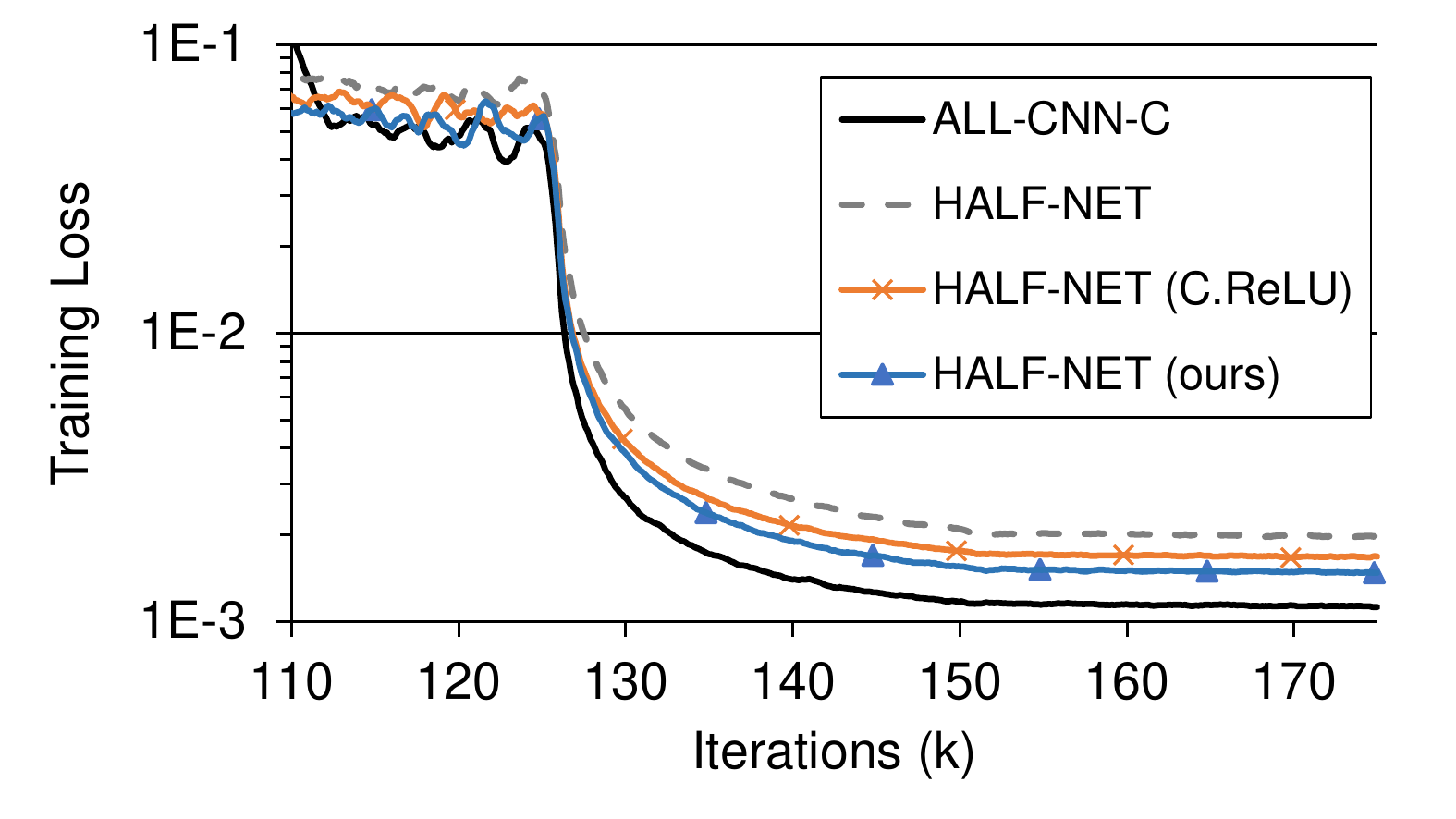}
    \fi
    \captionof{figure}{Training loss on CIFAR-10. Loss values are moving-averaged for better visualization.}
    \label{fig:crelu_trloss}
\end{minipage}
\hspace{0.5cm}
\begin{minipage}{0.55\textwidth}
    \centering
    \captionof{table}{Error rates on CIFAR-10. All the results are based on our training and test without data augmentation.}
    \small
    \begin{tabular}{lrr}
    Model           & Error (\%)    & Cost (MMACs) \\
    \hline
    ALL-CNN-C (our reproduction) & 9.83          & 270 \\
    HALF-CNN  & 10.91         & 72 \\
    HALF-CNN-CReLU & 9.99 & 140 \\
    HALF-CNN-mCReLU (ours) & 9.84 & 140 \\
    \end{tabular}
    \label{table:crelu_testerror}
\end{minipage}
\end{figure}

\paragraph{Modified C.ReLU}
C.ReLU \cite{ShangW2016icml} is motivated from an interesting observation of intermediate activation patterns in CNNs. In the early stage, output nodes tend to be ``paired'' such that one node's activation is the opposite side of another's.
From this observation, C.ReLU can double the number of output channels by simply concatenating negated outputs before applying ReLU.

The original design of C.ReLU enforces a shared bias between two negatively correlated outputs while the observations are about weight matrices only. We add a separated bias layer so that two correlated filters can have different bias values\ifemdnn.\fi
\ifemdnn
When it is tested with ALL-CNN-C network \cite{springenberg2015striving} on CIFAR-10, our modified C.ReLU shows lower training loss (Figure \ref{fig:crelu_trloss}) and better test accuracy (Table \ref{table:crelu_testerror}) compared to the original work.
\fi

\paragraph{Inception structure}

For object detection tasks, Inception has neither been widely applied nor been verified for its effectiveness.
We have found that Inception can be one of the most cost-effective building blocks for capturing both small and large objects in an input image.

To learn visual patterns for large objects, output features of CNNs should correspond to sufficiently large receptive fields, which can be easily fulfilled by stacking up convolutions of 3x3 or larger kernels.
On the other hand, in capturing small-sized objects, output features do not necessarily need to have large receptive fields, and a series of large kernels may lead to redundant parameters and computations. 1x1 convolution in Inception structure prevents the growth of receptive fields in some paths of the network, and therefore, can reduce those redundancies.


\paragraph{Deep network training}

It is widely accepted that as the network goes deeper and deeper, the training of the network becomes more troublesome. We solve this issue by adopting residual structures with pre-activation \cite{he2016identity} and batch normalization \cite{SzegedyC2015icml}. Unlike the original work, we add residual connections onto inception layers as well. 
\ifemdnn

We also implement our own policy to control the learning rate dynamically based on ``plateau detection''. After measuring the moving average of loss, if the minimum loss is not updated for a certain number of iterations, we call it {\em on-plateau}. Whenever plateau is detected, the learning rate is decreased by a constant factor. In experiments, our learning rate policy gave a notable gain of accuracy.
\fi

\paragraph{Overall design}

\begin{figure}[t]
\centerline{
\includegraphics[height=0.15\paperheight]{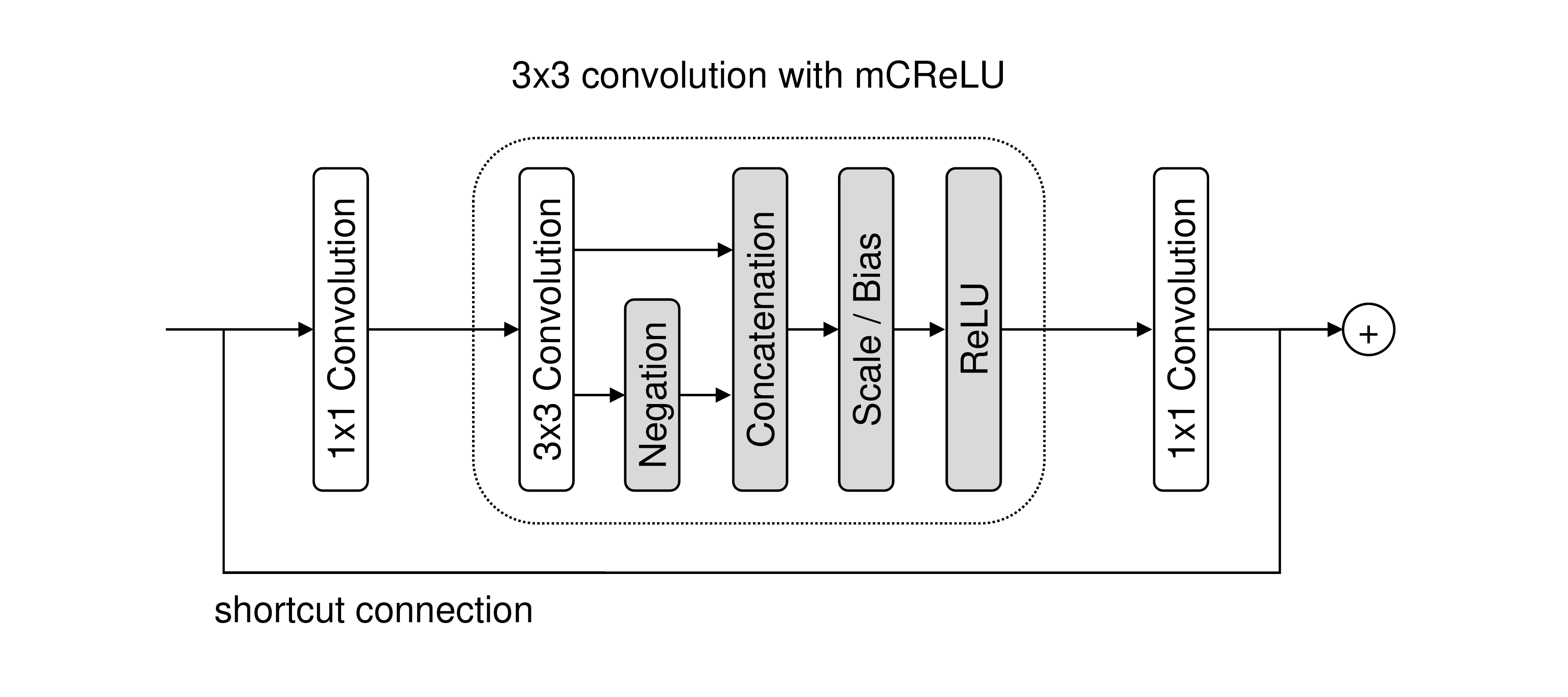}
\hspace{-1cm}
\includegraphics[height=0.15\paperheight]{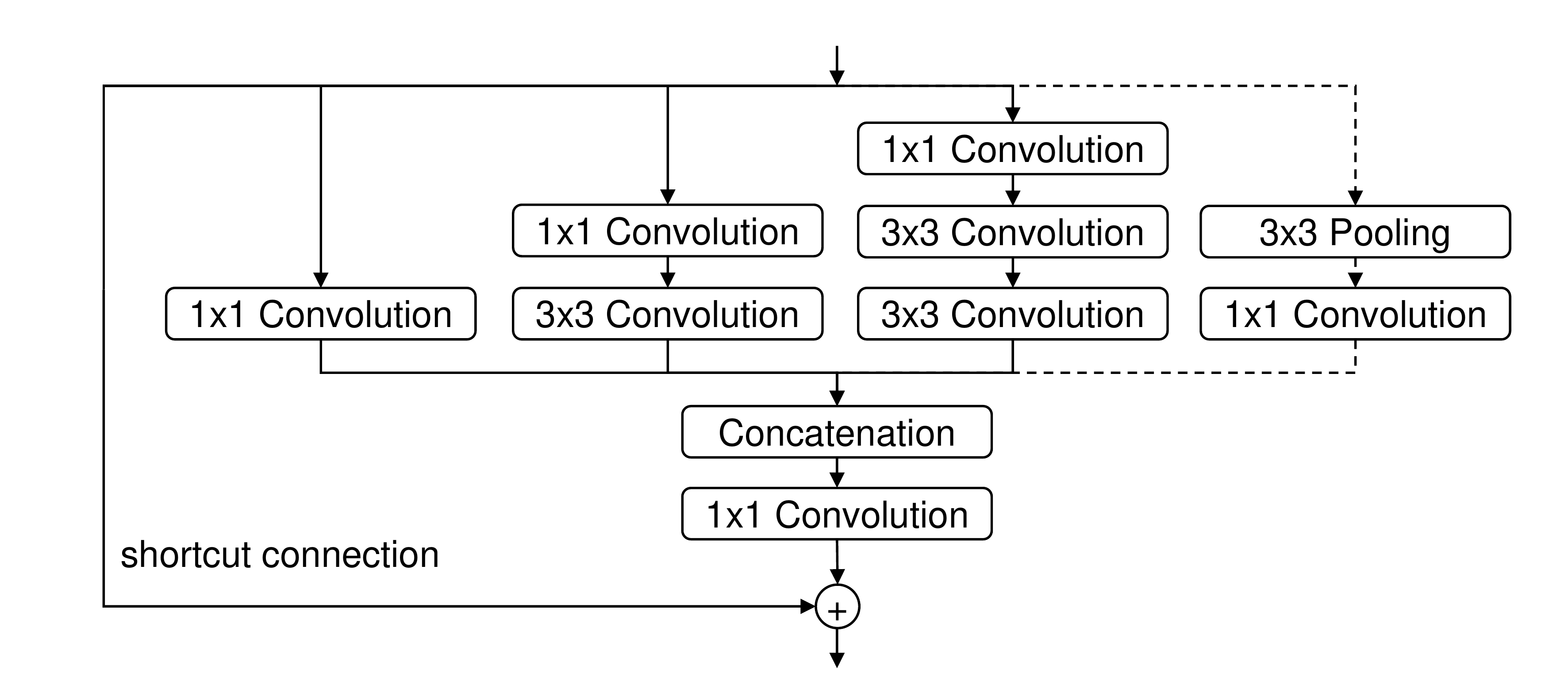}
}
\caption{Main building blocks of {\sc PVANet}. Every convolutional layer in these building blocks has its corresponding activation layers, a BatchNorm and a ReLU layer. However, they are not drawn here for simplicity. (Left) mCReLU building block (Right) Inception building block.}
\label{fig:building_block}
\end{figure}

Table \ref{table:pvanet_structure} shows the feature extraction network of {\sc PVANet}. In the early stage (conv1\_1, ..., conv3\_4) of the network, we adopt ``bottleneck'' building blocks\cite{HeK2016cvpr} in order to reduce the input dimensions of 3x3 kernels without jeopardizing overall representation capacity, and then we apply modified C.ReLU after 7x7 and 3x3 convolutional layers. The latter part of the network consists of Inception structures without the modified C.ReLU. In our Inception blocks, a 5x5 convolution is substituted by two consecutive 3x3 convolutional layers with an activation layer between them. With this feature extraction network, we can create an efficient network for object detection. Figure \ref{fig:building_block} shows the designs of two main building blocks of our network structure.

\subsection{Object detection network}
\label{subsec:frcnn}

\begin{table}
\caption{
The structure of the feature extraction network.
Theoretical computational cost is given as the number of multiplications and accumulations (MAC), assuming that the input image size is 1056x640.
{\bf KxK mCReLU} refers to a sequence of ``1x1 - KxK - 1x1'' convolutional layers where KxK is a block with the modified C.ReLU and describes the number of output channels of each convolutional layer. conv1\_1 has no 1x1 conv layer.
\vspace{0.4cm}
}
\hspace{-1.3cm}
\scriptsize
\begin{tabular}{c|c|cc|c|c|ccccc|rr}
Name & Type & Stride & Output & Residual     & mCReLU       & \multicolumn{5}{|c|}{Inception}         & \# params & MAC\\
     &      &        & size   &              & \#1x1-KxK-1x1 & \#1x1 & \#3x3 & \#5x5 & \#pool & \#out  &           & \\
\hline
conv1\_1 & 7x7 mCReLU    & 2 & 528x320x32 &    & X-16-X    &    & & & & & 2.4K & 397M\\
pool1\_1 & 3x3 max-pool & 2 & 264x160x32 &    &           &    & & & & &      &   \\
conv2\_1 & 3x3 mCReLU &   & 264x160x64 & O  & 24-24-64  &    & & & & & 11K & 468M\\
conv2\_2 & 3x3 mCReLU &   & 264x160x64 & O  & 24-24-64  &    & & & & & 9.8K & 414M\\
conv2\_3 & 3x3 mCReLU &   & 264x160x64 & O  & 24-24-64  &    & & & & & 9.8K & 414M\\
conv3\_1 & 3x3 mCReLU & 2 & 132x80x128 & O  & 48-48-128 &    & & & & & 44K & 468M\\
conv3\_2 & 3x3 mCReLU &   & 132x80x128 & O  & 48-48-128 &    & & & & & 39K & 414M\\
conv3\_3 & 3x3 mCReLU &   & 132x80x128 & O  & 48-48-128 &    & & & & & 39K & 414M\\
conv3\_4 & 3x3 mCReLU &   & 132x80x128 & O  & 48-48-128 &    & & & & & 39K & 414M\\
conv4\_1 & Inception   & 2 & 66x40x256 & O  &           & 64 & 48-128 & 24-48-48 & 128 & 256 & 247K & 653M\\
conv4\_2 & Inception   &   & 66x40x256 & O  &           & 64 & 64-128 & 24-48-48 &     & 256 & 205K & 542M\\
conv4\_3 & Inception   &   & 66x40x256 & O  &           & 64 & 64-128 & 24-48-48 &     & 256 & 205K & 542M\\
conv4\_4 & Inception   &   & 66x40x256 & O  &           & 64 & 64-128 & 24-48-48 &     & 256 & 205K & 542M\\
conv5\_1 & Inception   & 2 & 33x20x384 & O  &           & 64 & 96-192 & 32-64-64 & 128 & 384 & 573K & 378M\\
conv5\_2 & Inception   &   & 33x20x384 & O  &           & 64 & 96-192 & 32-64-64 &     & 384 & 418K & 276M\\
conv5\_3 & Inception   &   & 33x20x384 & O  &           & 64 & 96-192 & 32-64-64 &     & 384 & 418K & 276M\\
conv5\_4 & Inception   &   & 33x20x384 & O  &           & 64 & 96-192 & 32-64-64 &     & 384 & 418K & 276M\\
\hline
downscale & 3x3 max-pool & 2 & 66x40x128 &    &           &    & & & & &      &   \\
upscale   & 4x4 deconv   & 2 & 66x40x384 &    &           &    & & & & & 6.2K & 16M\\
concat    & concat       &   & 66x40x768 &    &           &    & & & & &      &   \\
convf     & 1x1 conv     &   & 66x40x512 &    &           &    & & & & & 393K & 1038M\\
\hline
Total     &              &   &           &    &           &    & & & & & 3282K & 7942M\\
\end{tabular}

\label{table:pvanet_structure}
\end{table}

\begin{figure}[t]
\centerline{
\includegraphics[width=1.2\columnwidth]{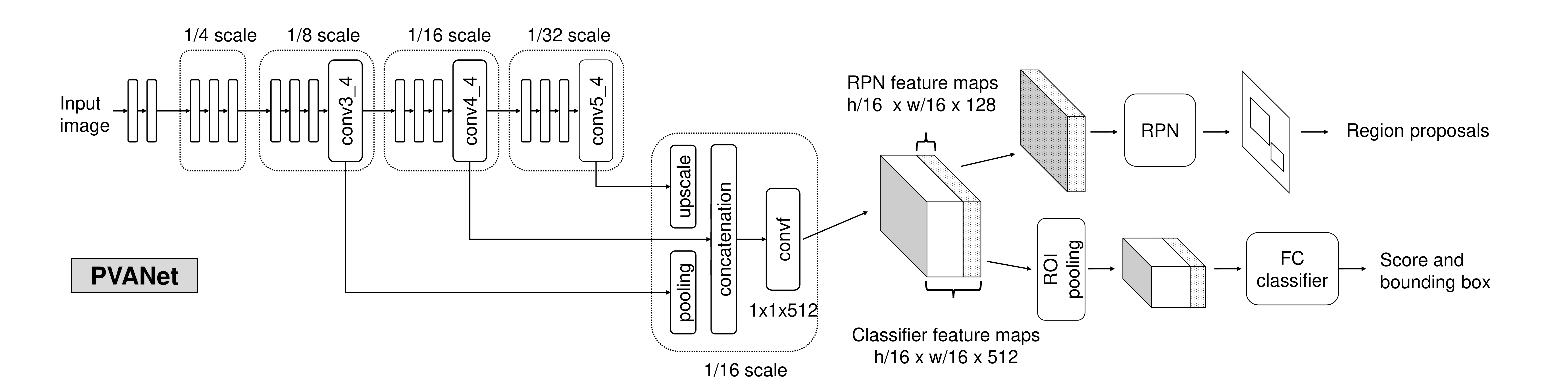}
}
\caption{The structure of {\sc PVANet} detection network.
}
\label{fig:faster_rcnn}
\end{figure}

Figure \ref{fig:faster_rcnn} shows the structure of {\sc PVANet} detection network. We basically follow the method of Faster R-CNN\cite{RenS2015nips}, but we introduce some modifications specialized for object detection. In this section, we describe the design of the detection network.

\paragraph{Hyper-feature concatenation}

Multi-scale representation and its combination are proven to be effective in many recent deep learning tasks \cite{KongT2016cvpr,BellS2016cvpr,HariharanB2015cvpr}.
Combining fine-grained details with highly abstracted information in the feature extraction layer helps the following region proposal network and classification network detect objects of different scales.
However, since the direct concatenation of all abstraction layers may produce redundant information with much higher compute requirement, we need to design the number of different abstraction layers and the layer numbers of abstraction carefully.

Our design choice is not different from the observation from ION \cite{BellS2016cvpr} and HyperNet \cite{KongT2016cvpr}, 
which combines 1) the last layer and 2) two intermediate layers whose scales are 2x and 4x of the last layer, respectively.
We choose the middle-sized layer as a reference scale (= 2x), and concatenate the 4x-scaled layer and the last layer with down-scaling (pooling) and up-scaling (linear interpolation), respectively. The concatenated features are combined by a 1x1x512 convolutional layer.

\paragraph{Towards a more efficient detection network}

In our experiments, we have found that feature inputs to the Region Proposal Network (RPN) does not need to be as deep as the inputs to the fully connected classifiers. Thanks to this observation, we feed only the first 128 channels in 'convf' into the RPN. This helps to reduce the computational costs by 1.4 GMAC without damaging its accuracy. The RPN in our structure consists of one 3x3x384 convolutional layer followed by two prediction layers for scores and bounding box regressions. Unlike the original Faster R-CNN \cite{RenS2015nips}, our RPN uses 42 anchors of 6 scales (32, 48, 80, 144, 256, 512) and 7 aspect ratios (0.333, 0.5, 0.667, 1.0, 1.5, 2.0, 3.0).

The classification network takes all 512 channels from 'convf'. For each ROI, 6x6x512 tensor is generated by ROI pooling, and then passed through a sequence of fully-connected layers of ``4096 - 4096 - (21+84)'' output nodes.\footnote{For 20-class object detection, R-CNN produces 21 predicted scores (20 classes + 1 background) and 21x4 predicted values of 21 bounding boxes.}
Note that our classification network is intentionally composed of fully-connected(FC) layers rather than fully convolutional layers. FC layers can be compressed easily without a significant accuracy drop \cite{RenS2015nips} and provide  possibility to balance between computational cost and accuracy of a network.



\section{Experimental results}
\label{sec:experiments}

\subsection{ImageNet Pre-training}
{\sc PVANet} is pre-trained with ImageNet 2012 classification dataset.
During pre-training, all images are resized into 256, 384 and 512. The network inputs are randomly cropped 192x192 patches due to the limitation in the GPU memory. The learning rate is initially set to 0.1, and then decreased by a factor of $1/\sqrt{10} \approx 0.3165$ whenever a plateau is detected. Pre-training terminates when the learning rate drops below $1e-4$, which usually requires about 2M iterations.

To evaluate the performance of our pre-trained network, we re-train the last three fully connected layers (fc6, fc7, fc8) with 224x224 input patches. Table \ref{table:ilsvrc2012} shows the accuracy of our network as well as others'. Thanks to its efficient network structure and training schemes, {\sc PVANet} shows a surprisingly competitive accuracy considering its computational cost. Its accuracy is even better than GoogLeNet\cite{SzegedyC2015cvpr}.

\begin{table}
\centering
\caption{Classification performance on the ImageNet 2012 validation set. Tested with single-scale, 10-crop evaluation. Shown VGG-16 10-crop results are reported by \cite{HeK2016cvpr}.
\vspace{0.4cm}
}
\begin{tabular}{lrrr}
Model       & top-1 err. (\%)   & top-5 err. (\%)   & Cost (GMAC) \\
\hline
AlexNet \cite{krizhevsky2012imagenet}     & 40.7             & 18.2            & 0.67 \\
VGG-16 \cite{simonyan2014very}     & 28.07             & 9.33             & 15.3 \\
GoogLeNet \cite{SzegedyC2015cvpr}   & -                 & 9.15              & 1.5 \\
ResNet-152 \cite{HeK2016cvpr}  & 21.43             & 5.71              & 11.3 \\
\hline
{\sc PVANet} & 27.66            & 8.84              & 0.6 \\
\end{tabular}
\label{table:ilsvrc2012}
\end{table}

\subsection{VOC2007 detection}

For the PASCAL VOC2007 detection task, the network is trained with the union set of MS COCO trainval, VOC2007 trainval and VOC2012 trainval and then fine-tuned with VOC2007 trainval and VOC2012 trainval. Training images are resized randomly so that shorter edges of inputs are between 416 and 864. All parameters are set as in the original work \cite{RenS2015nips} except for the number of proposal boxes before non-maximum suppression (NMS) ($= 12000$), the NMS threshold ($= 0.4$) and the input size ($= 640$). All evaluations were done on Intel i7-6700K CPU with a single core and NVIDIA Titan X GPU.

Table \ref{table:voc2007} shows the object recall and accuracy of our models in different configurations. Thanks to Inception structure and multi-scale features, our RPN generates initial proposals very accurately. It can capture almost 99\% of the target objects with only 200 proposals. Since the results imply that more than 200 proposals do not give notable benefits to object recall and detection accuracy, we fix the number of proposals to 200 in other experiments. 
The overall detection accuracy of plain {\sc PVANet} in VOC2007 reaches 84.4\% mean AP. When bounding box voting \cite{GidarisS2015iccv} is applied, the performance increases by 0.5\% mean AP. Unlike the original work, we do not apply an iterative localization and penalize object scores if there are less than 5 overlapped detections in order to suppress false alarms. We have found that the voting scheme works well even without an iterative localization.

The classification sub-network in {\sc PVANet} consists of fully-connected layers which can be compressed easily without a significant drop of accuracy \cite{GirshickR2015iccv}. When the fully-connected layers of ``4096 - 4096'' are compressed into ``512 - 4096 - 512 - 4096'' and then fine-tuned, the compressed network can run in 31.3 FPS with only 0.5\% accuracy drop.

\begin{table}
\centering
\caption{Performance on VOC2007-test.
{\sc PVANet+} denotes that bounding-box voting is applied, and `compressed' denotes that fully-connected layers are compressed.
\vspace{0.4cm}
}
\begin{tabular}{l|rrr|rr}
Model         & Proposals & Recall (\%) & mAP (\%)   & Time (ms) & FPS\\
\hline
{\sc PVANet}  & 300       & 99.2        & 84.4       & 48.5      & 20.6\\
              & 200       & 98.8        & 84.4       & 42.2      & 23.7\\
              & 100       & 97.7        & 84.0       & 40.0      & 25.0\\
              & 50        & 95.9        & 83.2       & 26.8      & 37.3\\
\hline
{\sc PVANet+} & 200       & 98.8        & {\bf 84.9} & 46.1      & 21.7\\
{\sc PVANet+} compressed & 200 & 98.8 & 84.4 & 31.9      & 31.3\\
\end{tabular}
\label{table:voc2007}
\end{table}

\subsection{VOC2012 detection}
For the PASCAL VOC2012 detection task, we use the same settings with VOC2007 except that we fine-tune the network with VOC2007 trainval, test and VOC2012 trainval.

Table \ref{table:voc2012} summarizes the comparisons between {\sc PVANet+} and some state-of-the-art networks \cite{HeK2016cvpr,RenS2015nips,Jif2016nips,liu15ssd} from the PASCAL VOC2012 leaderboard.\footnote{\scriptsize\url{http://host.robots.ox.ac.uk:8080/leaderboard/displaylb.php?challengeid=11\&compid=4}\label{footnote voc}}
Our {\sc PVANet+} has achieved the 4th place on the leaderboard as of the time of submission, and the network shows 84.2\% mAP, which is significant considering its computational cost. Our network outperforms ``Faster R-CNN + ResNet-101'' by 0.4\% mAP.

It is worthwhile to mention that other top-performers are based on ``ResNet-101'' or ``VGG-16'' which is much heavier than {\sc PVANet}. Moreover, most of them, except for ``SSD512'', utilize several time-consuming techniques such as global contexts, multi-scale testing or ensembles. Therefore, we expect that other top-performers are slower than our design by an order of magnitude (or more). Among the networks performing over 80\% mAP, {\sc PVANet+} is the only network running in $\leq 50$ms. Taking its accuracy and computational cost into account, {\sc PVANet+} is the most efficient network on the leaderboard.

It is also worth comparing ours with ``R-FCN'' and ``SSD512''. They introduced novel detection structures to reduce computational cost without modifying the base networks, while we mainly focus on designing an efficient feature-extraction network. Therefore, their methodologies can be easily integrated with {\sc PVANet} and further reduce its computational cost.

\begin{table}
\small
\centering
\caption{Performance on VOC2012-test of ours and some state-of-the-arts \cite{HeK2016cvpr,RenS2015nips,Jif2016nips,liu15ssd}.
Competitors' MAC are estimated from their Caffe prototxt files which are publicly available. Here, we assume that all competitors take an 1000x600 image and the number of proposals is 200 except for ``SSD 512'' which takes a 512x512 image.
Competitors' runtime performances are from \cite{RenS2015nips, Jif2016nips} and the public VOC leaderboard\textsuperscript{\ref{footnote voc}} while we projected the original values with an assumption that NVIDIA Titan X is 1.5x faster than NVIDIA K40.
\vspace{0.4cm}
}
\begin{tabular}{l|rrrr|rr|c}
Model                     & \multicolumn{4}{|c|}{Computation cost (GMAC)} & \multicolumn{2}{|c|}{Running time} & mAP\\
                          & Shared CNN & RPN  & Classifier & Total       & ms   & {\scriptsize$\times$({\sc PVANet})}     & (\%)\\
\hline
{\sc PVANet+}            & 7.9        & 1.4  & 18.5      & 27.8                      & 46   & 1.0                 & 84.2\\
{\sc PVANet+} compressed & 7.9        & 1.4  & 3.2       & 12.5                      & 32   & 0.7                 & 83.7\\
\hline
Faster R-CNN + ResNet-101               & 80.5       & N/A  & 125.9     & \textgreater 206.4        & 2240 & 48.6                & 83.8\\
Faster R-CNN + VGG-16                   & 183.2      & 2.7  & 18.5      & 204.4                     & 110  & 2.4                 & 75.9\\
R-FCN + ResNet-101                      & 122.9      & 0    & 0         & 122.9                     & 133  & 2.9                 & 82.0\\
SSD512 (VGG16)                          & 86.7       & 0    & N/A       & \textgreater 86.7         & 53   & 1.15                & 82.2
\end{tabular}
\label{table:voc2012}
\end{table}

\section{Conclusions}

In this paper, we show that the current networks are highly redundant and that we can design a thin and light network which is capable of complex vision tasks.
Elaborate adoptions and combinations of recent technical innovations on deep learning make it possible for us to design a network to maximize the computational efficiency.
Even though the proposed network is designed for object detection, we believe that our design principle is widely applicable to other tasks such as face recognition and semantic analysis.

Our network design is completely independent of network compression and quantization.
All kinds of recent compression and quantization techniques are applicable to our network as well to further increase the actual performance in real applications.
As an example, we show that a simple technique like truncated SVD could achieve a notable improvement in the runtime performance based on our network.

\renewcommand\refname{\subsubsection*{References}}
\small{
\bibliographystyle{unsrt}
\bibliography{khk}
}

\end{document}